\documentclass[letterpaper, 10 pt, conference]{ieeeconf}  

\IEEEoverridecommandlockouts                              

\usepackage[utf8]{inputenc}
\usepackage{graphicx}
\usepackage{epsfig}
\usepackage{mathptmx}
\usepackage{booktabs}  
\usepackage{times}
\usepackage{epstopdf}
\usepackage{amsmath}
\usepackage{amssymb}
\usepackage{mathrsfs}
\usepackage{subfigure}
\usepackage{url}
\usepackage[colorlinks,linkcolor=red]{hyperref}
\usepackage{hyperref}
\usepackage[hyphenbreaks]{breakurl}
\usepackage{multirow}

\title{\LARGE \bf
	A Comparative Study of Deep Reinforcement Learning-based Transferable Energy Management Strategies for Hybrid Electric Vehicles
}

\author{Jingyi Xu$^{1}$, Zirui Li$^{2}$, Li Gao$^{1}$, Junyi Ma$^{1}$, Qi Liu$^{1}$ and Yanan Zhao$^{1,*}$
	\thanks{This work was supported by the National Key R\&D Program of China under Grant 2018YFB0105205-02, Grant 2017YFC0804803 and Grant 2017YFC0804808.}
	\thanks{$^1$ Jingyi Xu, Li Gao, Junyi Ma, Qi Liu and Yanan Zhao are with School of Mechanical Engineering, Beijing Institute of Technology, 100081 Beijing, China
	}%
	\thanks{$^2$ Zirui Li is with the Department of Transport and Planning, Delft University of Technology. Delft 2628 CD, The Netherlands and is also with the School of Mechanical Engineering, Beijing Institute of Technology, 100081 Beijing, China
	} %
	\thanks{* Corresponding author: Y. Zhao and Z. Li}
}

\begin{document}

	\maketitle
	\thispagestyle{empty}
	\pagestyle{empty}

	\begin{abstract}
		
		The deep reinforcement learning-based energy management strategies (EMS) have become a promising solution for hybrid electric vehicles (HEVs). When driving cycles are changed, the neural network will be retrained, which is a time-consuming and laborious task. A more efficient way of choosing EMS is to combine deep reinforcement learning (DRL) with transfer learning, which can transfer knowledge of one domain to the other new domain, making the network of the new domain reach convergence values quickly. Different exploration methods of DRL, including adding action space noise and parameter space noise, are compared against each other in the transfer learning process in this work. Results indicate that the network added parameter space noise is more stable and faster convergent than the others. In conclusion, the best exploration method for transferable EMS is to add noise in the parameter space, while the combination of action space noise and parameter space noise generally performs poorly. Our code is available at \url{https://github.com/BIT-XJY/RL-based-Transferable-EMS.git}.
		
	\end{abstract}

	\section{Introduction}
	
	Hybrid electric vehicles (HEVs) are currently important carriers of self-driving technology \cite{i1}. HEVs involve two or more energy sources. Thus there is a considerable need for energy management strategies (EMS) to distribute power supplements among several power sources to improve energy efficiency and reduce emissions \cite{i2}. There are mainly three types of EMS for HEVs: rule-based methods, optimization-based techniques, and learning-based approaches \cite{i3}.
	
	The rule-based approach is the most common method to achieve real-time control of HEVs, the effectiveness of which depends on the intuition and experience of engineers \cite{i6}. To reduce the reliance on professional engineers, the optimization-based method is introduced, using optimization algorithms to solve for optimal or sub-optimal solutions in the feasible domain to obtain better fuel economy \cite{i7}. According to different optimal control objectives and algorithms, optimization-based EMS can be divided into global optimization and real-time optimization approaches. The classical optimization-based methods include linear programming algorithm (LP) \cite{i8}, dynamic programming algorithm (DP) \cite{i10,i11}, equivalent fuel consumption minimization strategy (ECMS) \cite{i12,i13}, model predictive control (MPC) \cite{i14}, etc. The above methods improve the real-time performance and fuel economy of EMS to some extent, but they have more computational cost than rule-based methods \cite{i15}.

	With the rapid development of machine learning in recent years, the learning-based energy management method has become a promising solution for HEVs. Current studies mainly focus on deep reinforcement learning (DRL) based EMS due to their strong learning ability, where the EMS problem is modeled as a Markov Decision Process (MDP). The optimal solution for EMS can be learned through the interaction between agents and the environment.  \cite{i16} used deep Q-learning network (DQN) algorithm for energy management, which solved the dimensional catastrophe problem. Based on this, \cite{i17} compared double deep Q-learning with DQN for energy management of plug-in hybrid vehicles and demonstrated advantages of the former in terms of convergence and fuel economy. \cite{i18} showed that the energy management policy based on deep deterministic policy gradient (DDPG) algorithm has a strong characterization capability of deep neural networks and can improve fuel economy significantly. In addition, \cite{i19} indicated that asynchronous advantage actor-critic (A3C) and distributed proximal policy optimization (DPPO) improved the learning efficiency. 
	
	Although DRL-based methods have made a significant breakthrough, their limitations are the long training time for an agent to learn the optimal solution through trial-and-error interactions with the environment \cite{i20}. Besides, the training process must be repeated even when encountering a new but similar task. Therefore, some works have combined transfer learning with DRL to improve the training efficiency between similar tasks. \cite{i21} combined proximal policy optimization (PPO) and transfer learning to effectively reduce time consumption and guarantee control performance. \cite{i22} combined DDPG and transfer learning to derive an adaptive energy management controller for hybrid tracked vehicles. Results show that this method has the potential to be applied in real-world environments. \cite{i23} incorporated transfer learning into DDPG-based EMS for HEVs to transfer knowledge among three types of HEVs that have apparently different structures.
	
	In DRL, the agent utilizes exploration methods to acquire knowledge about the environment, which may explore better actions. The main approach is to add different types of noise while selecting actions. Comparing the impact of different exploration methods on DRL is implemented by much previous work \cite{i24,i25}. However, there are few studies considering the effects of exploration methods on the combination of DRL and transfer learning, which improve the training efficiency of algorithms and reduce the computational cost.
	
	This work is a comparative study, which focuses on effects of different exploration methods of DDPG for transferable EMS. DDPG combines advantages of DQN and the actor-critic architecture, which leads to stability and high efficiency. Thus, DDPG is appropriate for evaluating the strategies from network parameters transferring. In this work, several types of noise are added to DDPG netwoks which are trained by multiple driving cycles. Then, training weights are saved to initialize a new DDPG network. The second training process is performed with noise to acquire the optimal transferable EMS.
	
	In sum, the main contributions of this work are: Different types of noise, i.e., action space noise and parameter space noise, are added to the DDPG algorithm to explore in actions selection. Parameters of networks with different exploration methods are used to initialize new networks. The methods of exploration that work best for DDPG-based EMS and suit the most for transfer learning in the training efficiency are given by the comparative study.
	
	The remainder of this work is organized as follows: Section \uppercase\expandafter{\romannumeral2} introduces the proposed method in comparing effects of different exploration approaches of DDPG-based EMS and the performance of the transferred new network; Section \uppercase\expandafter{\romannumeral3} details experiment results, and the conclusion is depicted in Section \uppercase\expandafter{\romannumeral4}.
	
	\section{Deep Reinforcement Learning-based Transferable Energy Management Strategies}
	
\begin{figure}[thpb]
		\centering
		\includegraphics[width=0.95\linewidth]{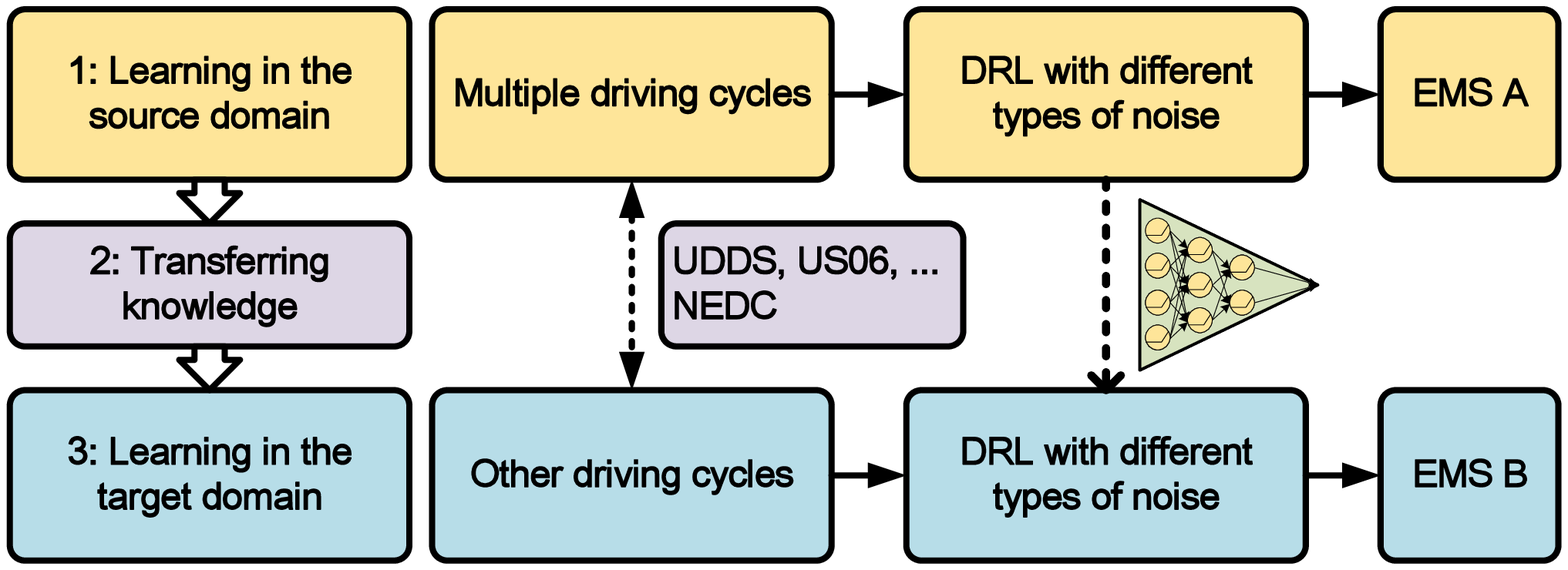}
		\caption{The sketch map of DRL-based transferable EMS.}
		\label{map}
	\end{figure}

	A DRL-based transferable EMS is used to evaluate the performance of different exploration methods. The sketch map of DRL-based transferable EMS is shown in Fig.\ref{map}. This section describes the HEV model, the DRL-based EMS formulation, different types of noise added to DRL networks, and the effects of transferred new domain networks using different kinds of noise. In this process, the effects of different types of noise for exploration in DDPG and deep transfer learning are compared in detail in Section \uppercase\expandafter{\romannumeral3}.
	
	\subsection{Hybrid Electric Vehicle Model}
	
	The EMS for Prius, one of the most classical HEVs, has been extensively studied \cite{m2}.
	
	\subsubsection{Prius configuration}
	
\begin{figure}[thpb]
		\centering
		\includegraphics[width=0.85\linewidth]{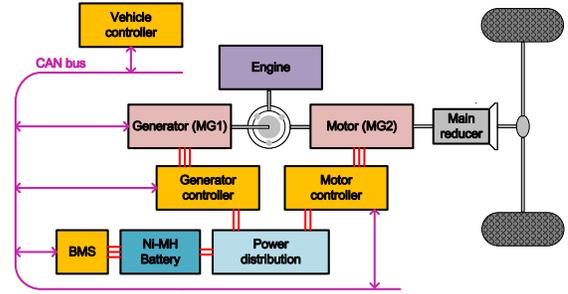} 
		\caption{Architecture of Prius powertrain.}
		\label{fig1}
	\end{figure}

	Prius is equipped with the Hybrid Synergy Drive system, which consists of an internal combustion engine ICE, an electric motor MG2, and a generator MG1. Prius is also equipped with a low-capacity nickel-metal hydride (Ni-MH) battery used to drive the motor and generator. These systems in Fig.\ref{fig1} are integrated with a power splitting planetary gear, which provides various power flow configurations for different operating.
	
	\subsubsection{Power request model}
	
	After building the Prius model, the vehicle power demand is calculated using the longitudinal force balance equation. The longitudinal force $F$ consists of rolling resistance $F_f$, aerodynamic drag $F_w$, gradient resistance $F_i$, and inertial force $F_a$ \cite{m3}:
	
	\begin{equation}\left \{\begin{aligned} & F = F_f + F_w + F_i + F_a \\ & F_f = mg \cdot f \\ & F_w = \frac{1}{2}\rho\cdot A_f \cdot C_D \cdot v^2 \\ & F_i = mg \cdot i \\ & F_a = m \cdot a \end{aligned}\right.\end{equation}
	
	\noindent where $m$ is the curb weight, $g$ is the gravitational constant, $f$ is the rolling friction coefficient, $\rho$ is the air density, $A_f$ is the fronted area, $C_d$ is the aerodynamic coefficient, $v$ is the speed in regard to a certain driving cycle, $i$ is the road slope (not considered in this paper), and $a$ is the acceleration.
	
	\subsubsection{Powertrain system model}
	
	The engine, the electric motor, and the generator of the Prius are modeled by their corresponding efficiency maps from bench tests. The Ni-MH battery is modeled by an equivalent circuit model ignoring temperature changing and battery aging:
	
	\begin{equation}\left \{\begin{aligned} & P(t) = I(t) \cdot V_{oc}(t) - R_0 \cdot I^2(t) \\ & I(t) = \frac{V_{oc}(t) - \sqrt{V^2_{oc}(t)-4 \cdot R_0 \cdot P(t)}}{2R_0} \\ & SoC(t) = \frac{Q_0 - \int_0^tI(t)dt}{Q} \end{aligned}\right.\end{equation}
	
	\noindent where $P$ is the output power, $I$ denotes the current, $V_{oc}$ is the open-circuit voltage, $R_0$ is the internal resistance, $SoC$ is the state of charge, $Q_0$ is the initial battery capacity, and $Q$ is the nominal battery capacity. Details on Prius parameters are shown in Table \ref{Prius}.
	
	\begin{table}[]
		\caption{Parameters of Prius}   
		\centering{    
			\begin{tabular}{ccc}
				\toprule
				\textbf{Components}           & \textbf{Parameters}               & \textbf{Values}  \\ \hline
				\multirow{2}{*}{Engine}       & Maximum power, $P_e$              & 56 kW            \\
				& Maximum torque, $T_e$             & 120 Nm           \\ \cline{2-3} 
				\multirow{2}{*}{Motor}        & Maximum power, $P_m$              & 50 kW            \\
				& Maximum torque, $T_m$             & 400 Nm           \\ \cline{2-3} 
				\multirow{2}{*}{Battery}      & Capacity, $Q$                     & 1.54 kWh         \\
				& Voltage, $V_oc$                   & 237 V            \\ \cline{2-3} 
				\multirow{5}{*}{Vehicle}      & Curb weight, $m$                  & 1449 kg          \\
				& Roll resistance coefficient, $f$  & 0.013            \\
				& Air resistance coefficient, $f_A$ & 0.26             \\
				& Frontal area, $A_f$               & 2.23 ${\rm m^2}$ \\
				& Wheel radius, $r$                 & 0.287 m          \\ \cline{2-3} 
				\multirow{2}{*}{Transmission} & Final gear ratio, $i_g$           & 3.93             \\
				& Characteristic parameter, $C$     & 2.6              \\ \bottomrule
			\end{tabular}
		} 
		\label{Prius}
	\end{table}
	
	\subsection{DRL Formulation}
	
	A DRL problem that satisfies the Markov property can generally be modeled in terms of the MDP, which can be characterized as $(\mathbf{S}, \mathbf{A}, \mathbf{P}, R, \gamma)$. $\mathbf{S}$ represents a set of state spaces. $\mathbf{A}$ is a set of action spaces. $\mathbf{P}$ denotes a state transition probability matrix. $R$ represents a reward function. $\gamma$ denotes a discount factor. 
	
\begin{figure}[thpb]
		\centering
		\includegraphics[width=1\linewidth]{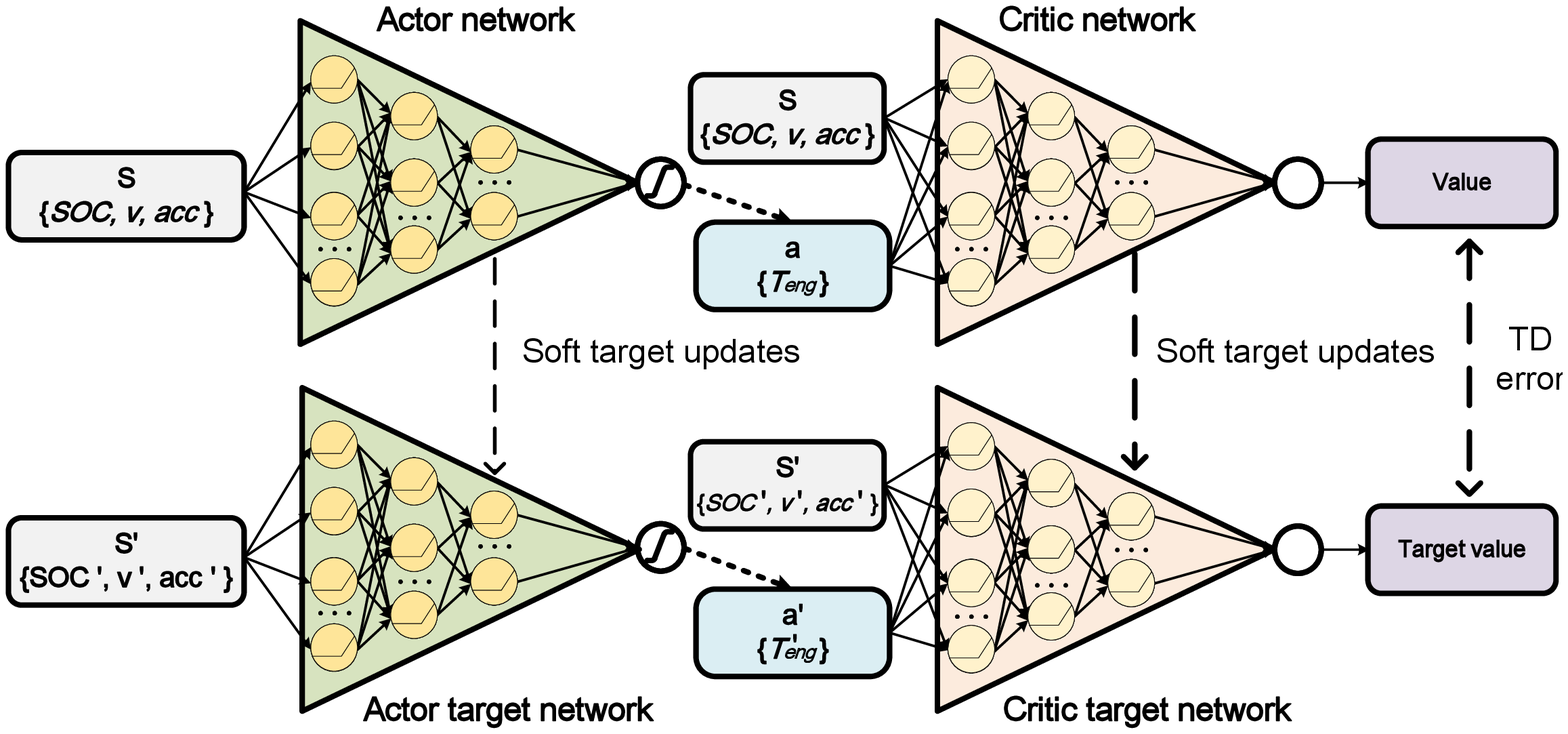}
		\caption{The architecture of DDPG.}
		\label{DDPG}
	\end{figure}

	DDPG is one of the most typical actor-critic DRL methods, which is an off-policy and model-free algorithm. As illustrated in Fig.\ref{DDPG}, DDPG has an actor network $\mu (\mathbf{s}|\theta ^ \mu)$, a critic network $Q (\mathbf{s},\mathbf{a}|\theta ^Q)$, an actor target network $\mu' (\mathbf{s}'|\theta ^ {\mu '})$, and a critic target network $Q' (\mathbf{s}',\mathbf{a}'|\theta ^{Q'})$. The actor target network has the same structure as the actor network, while the critic target network has the same structure as the critic network. $\mathbf{s}$ is the agent state as the input of actor network and critic network. $\mathbf{a}$ is the agent action as the output of actor network and the input of critic network. $\theta$ represents parameters of the corresponding network. $\mathbf{s}'$ and $\mathbf{a}'$ are defined in the same way.
	
	The DDPG algorithm is used to learn the optimal policy of Prius EMS in this work. The neural network has a pyramid-like architecture, with the number of neurons in hidden layers decreasing layer by layer. The optimal EMS utilizes the DDPG algorithm, which is trained with different driving cycles. 
	
	The DDPG-based EMS is formulated according to the following MDP.
	
	\subsubsection{State space, $\mathbf{S}$}
	The state of the system,
	
	\begin{equation}
		\mathbf{s} = \{SoC, v, acc\}
	\end{equation}
	
	\noindent which consists of $SoC$, the velocity of Prius $v$ and the acceleration $acc$.
	
	\subsubsection{Action space, $\mathbf{A}$}
	At each episode, the agent can select actions in continuous engine power $T_{eng}$:
	
	\begin{equation}
		\mathbf{a} = \{T_{eng}\}
	\end{equation}
	
	\subsubsection{Reward function, $R$}
	There are two aspects of the reward function of DDPG-based EMS, the energy consumption and the SoC sustaining. The multi-objective reward function is defined as:
	
	\begin{equation}
		r = -\{\alpha [fuel(t) + elec(t)] + \beta [SoC_{ref} - SoC(t)]^n\}
	\end{equation}
	
	\noindent where $\alpha$ is the weight of Prius consumption including the fuel consumption of engine $fuel$ and the electricity consumption of motor $elec$, $\beta$ is the weight of battery charge-sustaining, and $SoC_{ref}$ represents the SoC reference value. The goal of the reward function is to minimize the energy consumption and retain the battery SoC at an appropriate range. To control for variables, in the following comparison, $SoC_{ref}$ is selected as 0.6 according to the minimum charge-discharge internal resistance. $\alpha$ is selected as 1, $\beta$ is set to 350, and $n$ is set to 2, according to the previous work \cite{m3}.  
	
	\subsection{Transfer Learning}
	
	Traditional DRL algorithms are used to solve the problem with training and test data in the same domain. However, once the domain is changed, the network needs to be retrained, which is quite complex and time-consuming \cite{lu2019virtual, li2019transferable, gong2019comparative}. Transfer learning is extremely useful in solving this problem. When two domains are similar, network parameters can be stored and reused in the new one along with transfer learning approaches \cite{li2022personalized, li2020importance, lu2019transfer}.
	
	Given a source domain $M_s$ and a target domain $M_t$, transfer learning aims to learn an optimal policy $\pi ^*$ from $M_s$ for $M_t$. $M_s$ provides prior knowledge $D_s$ that is accessible for $M_t$. Thus, by leveraging the information from $D_s$, the target agent learns better and faster in $M_t$ \cite{t0}. 
	
	A network that specializes in obtaining source EMS is used in our work. Since driving cycles of $M_s$ and $M_t$ have the same feature space and are correlated with each other, source domain knowledge can be transferred to the novel, but relevant target domain \cite{t1}. The majority of parameters in the neural network are the same, and only parameters of the output layer should be retrained. Thus, both the source network and the target network use the same DDPG architecture shown in Fig.\ref{DDPG}, and the weights of the source network except for the last layer are used to initialize the target network that will be trained on new driving cycles. Further details about hyperparameters of DDPG in $M_s$ and $M_t$ are given in Table \ref{Hyperparameters}.
	
	In DDPG, the agent utilizes exploration to acquire knowledge about the environment and applies exploitation to select a control action based on current knowledge \cite{e1}. Thus, the coordination between exploitation and exploration is essential. The following parts of this subsection provide a description of exploration methods used in the DDPG-based transferable EMS.
	
	The primary purpose of exploration is to avoid local optimum for agent's behaviors \cite{i24}. Thus, to realize efficient and effective exploration, random noise is frequently added to perturb selected actions, which is mainly focused on in our work. Main approaches can be classified into two groups: adding noise in the action space and adding noise directly to agent's parameters.
	
	\subsubsection{Action space noise}
	When the agent selects actions using the actor network, the noise $\mathcal{N}$ is added to the action space. The final selected action $a_t$ at each step satisfies:
	
	\begin{equation}
		a_t = \mu (s|\theta ^ \mu) + \mathcal{N}
	\end{equation}
	
	Action space noise could be a simple Gaussian noise or a more advanced Ornstein-Uhlenbeck (OU) correlated noise process \cite{i25}. Gaussian noise satisfies $\mathcal{N} \sim N(0, \sigma ^2 I)$, where $\sigma ^2$ denotes variance and the expected value is set to 0. An OU process \cite{e4} can be used as a temporally correlated noise. Just like the Gaussian noise mentioned above, the expected value of OU noise $\mathcal{N} \sim OU(0, \sigma ^2)$ is set to 0, and the variance can be set to multiple values.
	
	\subsubsection{Parameter space noise}
	While adding noise in the action space to explore, there is no guarantee that the same action will be chosen in the same state each time, which can lead to inconsistent exploration. The parameter space noise solves this problem and directly perturbs actor network parameters to get a rich set of behaviors. The final selected action $a_t$ at each step satisfies:
	
	\begin{equation}
		\left \{\begin{aligned} & a_t = \mu (s|\Tilde{\theta} ^ \mu) \\ & \Tilde{\theta} = \theta + N(0, \sigma ^2 I) \end{aligned}\right.
	\end{equation}
	
	\begin{table}[]
		\caption{DDPG Hyperparameters}   
		\centering{    
			\begin{tabular}{ccc}
				\toprule
				Parameters                             & Source Domain                    & Target Domain \\ \hline
				Number of Episodes, $K$           & 1000                      & 300 \\
				Replay memory size, $M$                    & 50000                         & 50000 \\
				Learning rate of actor network, $lr_{a}$                        & 0.001                         & 0.0009 \\
				Learning rate of critic network, $lr_{c}$              & 0.01                         & 0.009 \\
				Discount factor, $\gamma$              & 0.9                         & 0.9 \\
				Target network update frequency, $\tau$              & 0.01                         & 0.01 \\
				Mini-batch size, $batch$              & 64                         & 64 \\ \bottomrule
			\end{tabular}
		} 
		\label{Hyperparameters}
	\end{table}
	
	\section{Experiments}
	
	The purpose of our work is to compare effects of different exploration methods of DDPG-based EMS and transferred new networks in terms of transfer efficiency. Driving cycles are selected for the source domain and the target domain, which are different but similar. Then, networks with different exploration methods are trained in the source domain, of which parameters are saved. Finally, the adaptation of target domain networks, of which parameters are initialized using saved weights, is evaluated.
	
	\subsection{Driving Cycles}
	
	In this work, driving cycles are all selected from standard data. Source tasks are performed over multiple cycles, including Urban Dynamometer Driving Schedule (UDDS) \cite{yang2021energy}, FTP75 \cite{m3}, etc. Target tasks are conducted on New European Driving Cycle (NEDC) \cite{liu2019heuristic}, which is different from driving cycles used in the source domain but similar. Using multiple driving cycles for training in the source domain improves the generalization ability of the trained model, which leads to better transfer results. A driving cycle with a high similarity to the source driving cycles is chosen for the target domain, since similarity is a necessary factor for transfer learning.
	
	\subsection{Training in the Source Domain}
	
	To ensure the validity of weights to be transferred, networks with different exploration methods are firstly trained on the source domain. Settings of networks with different types of noise are shown in Table \ref{source table}. By comparative studying, suitable networks, of which parameters are used to initiate weights of target networks, are selected according to the training results.
	
	\begin{table}[]
		\caption{Networks added different types of noise}
		\setlength{\tabcolsep}{0.2mm}{
			\begin{tabular}{ccc}
				\toprule
				\textbf{Space Added Noise}                 & \textbf{Noise Type}  & \textbf{Variance}        \\ \hline
				\multirow{2}{*}{Action space}              & Gaussian             & 0.02,0.03,0.04,0.05,\textbf{0.06} \\
				& OU                   & 0.08,\textbf{0.09},0.10,0.11,0.13 \\ \cline{2-3} 
				Parameter space                            & Gaussian             & \textbf{0.03},0.04                \\ \cline{2-3} 
				\multirow{2}{*}{Action \& parameter space} & Gaussian \& Gaussian & 0.06 \& 0.03             \\
				& OU \& Gaussian       & 0.09 \& 0.03             \\ \bottomrule
		\end{tabular}}
		\label{source table}
	\end{table}
	
	\begin{figure*}[thpb]
		\centering
		\subfigure[Action space noise with Gaussian]{
			\label{source_gs}
			\includegraphics[width=0.32\linewidth]{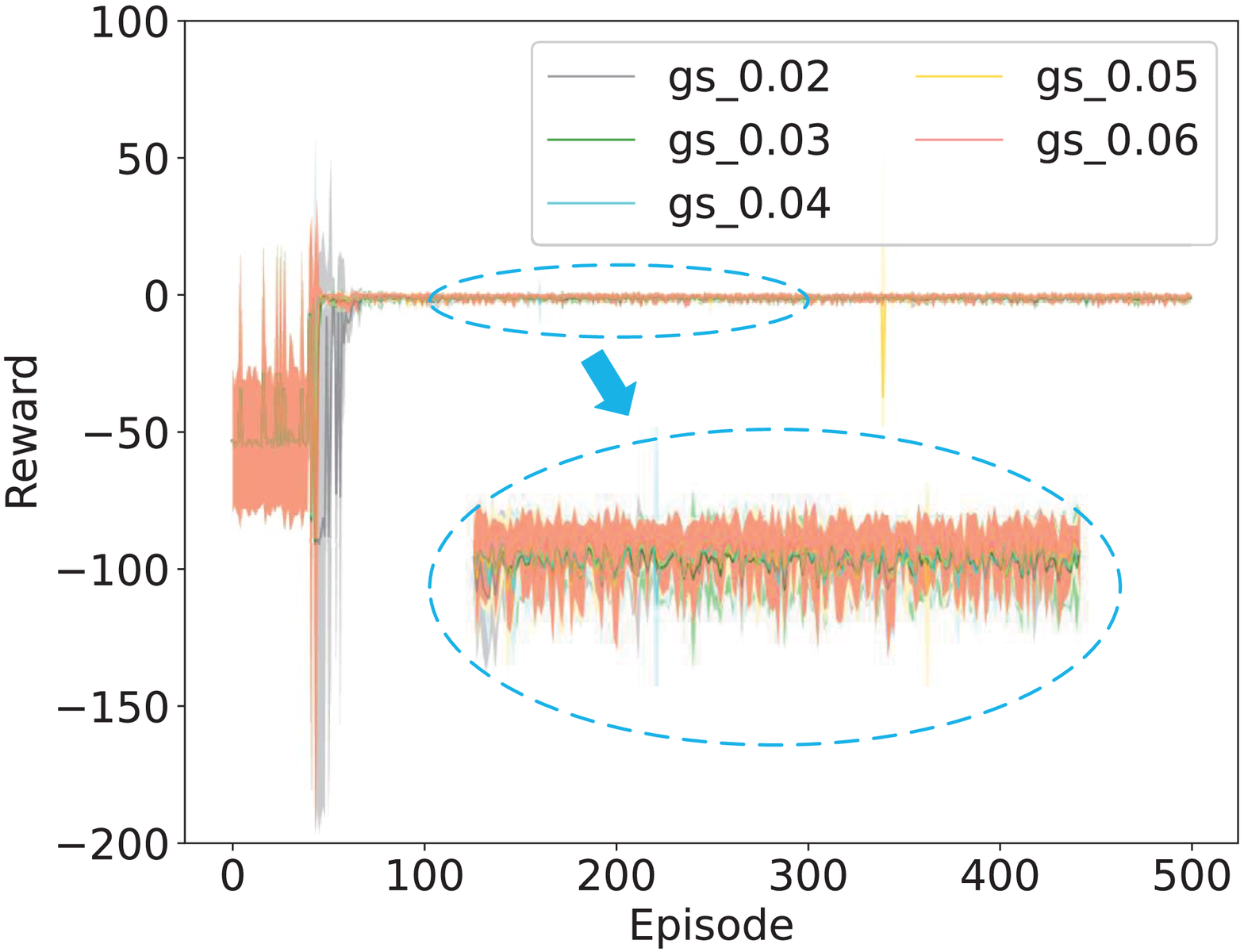}}\subfigure[Action space noise with OU]{
			\label{source_ou}
			\includegraphics[width=0.32\linewidth]{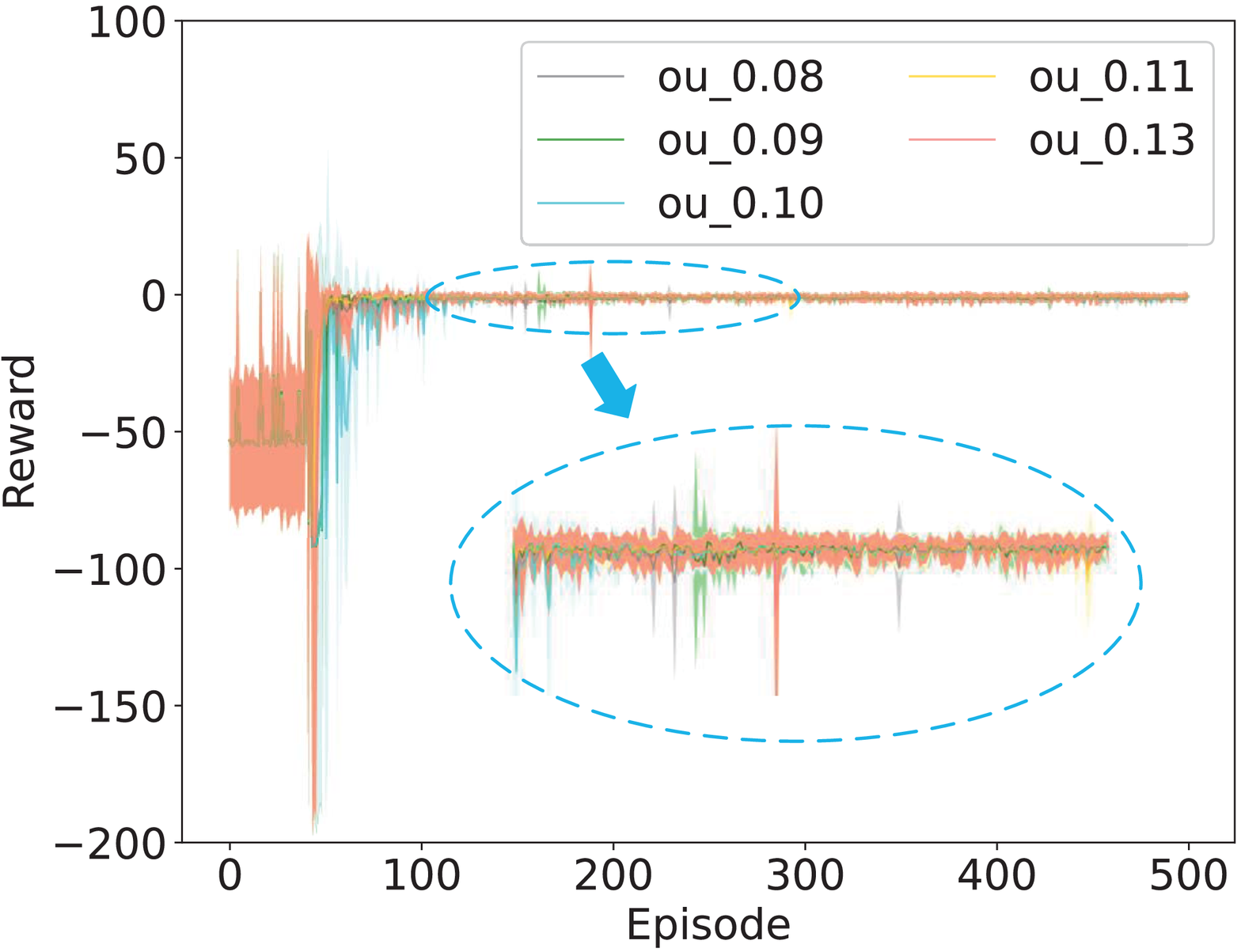}}\subfigure[Parameter space noise with Gaussian]{
			\label{source_param}
			\includegraphics[width=0.32\linewidth]{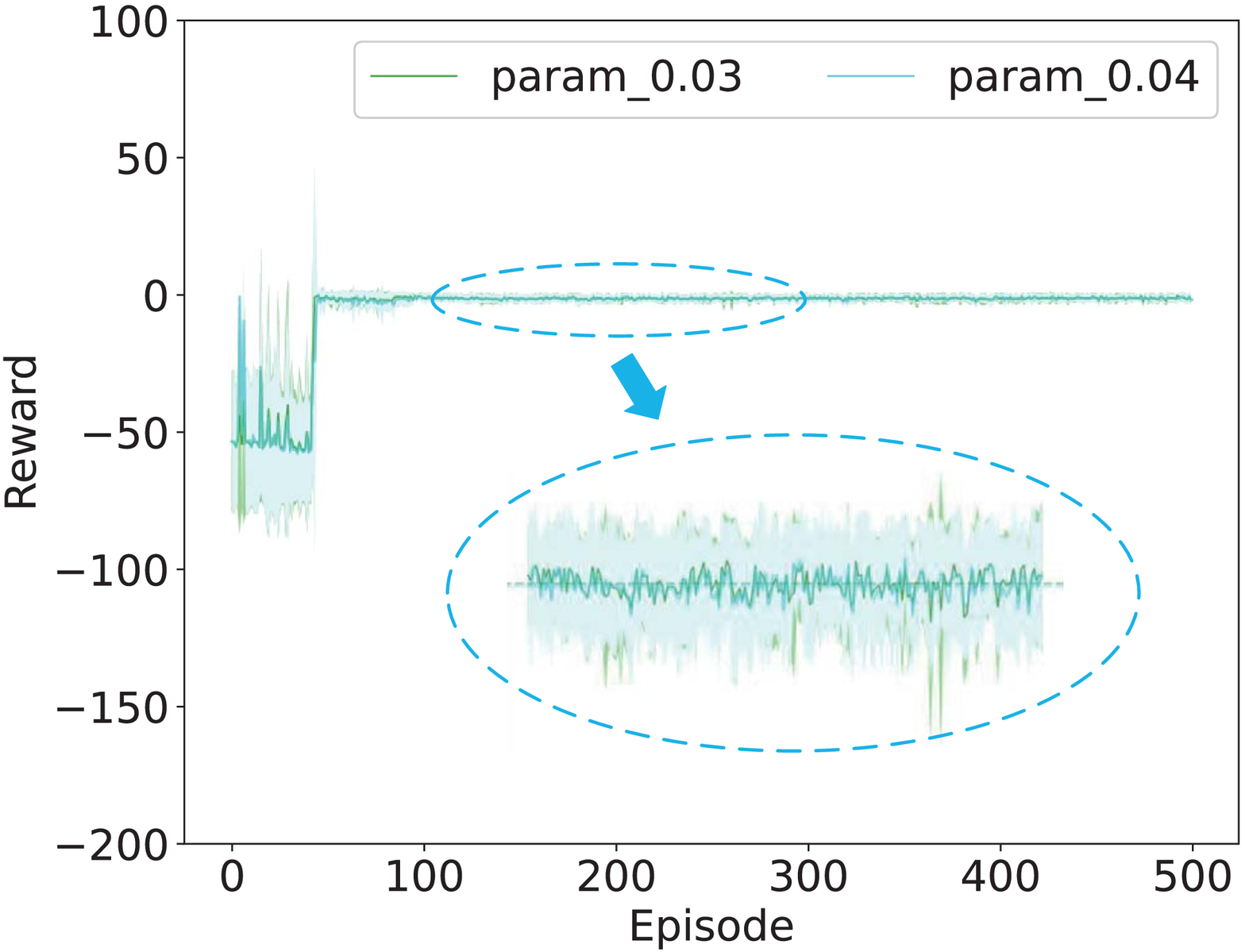}}
		\subfigure[Mixture noise]{
			\label{source_mix}
			\includegraphics[width=0.32\linewidth]{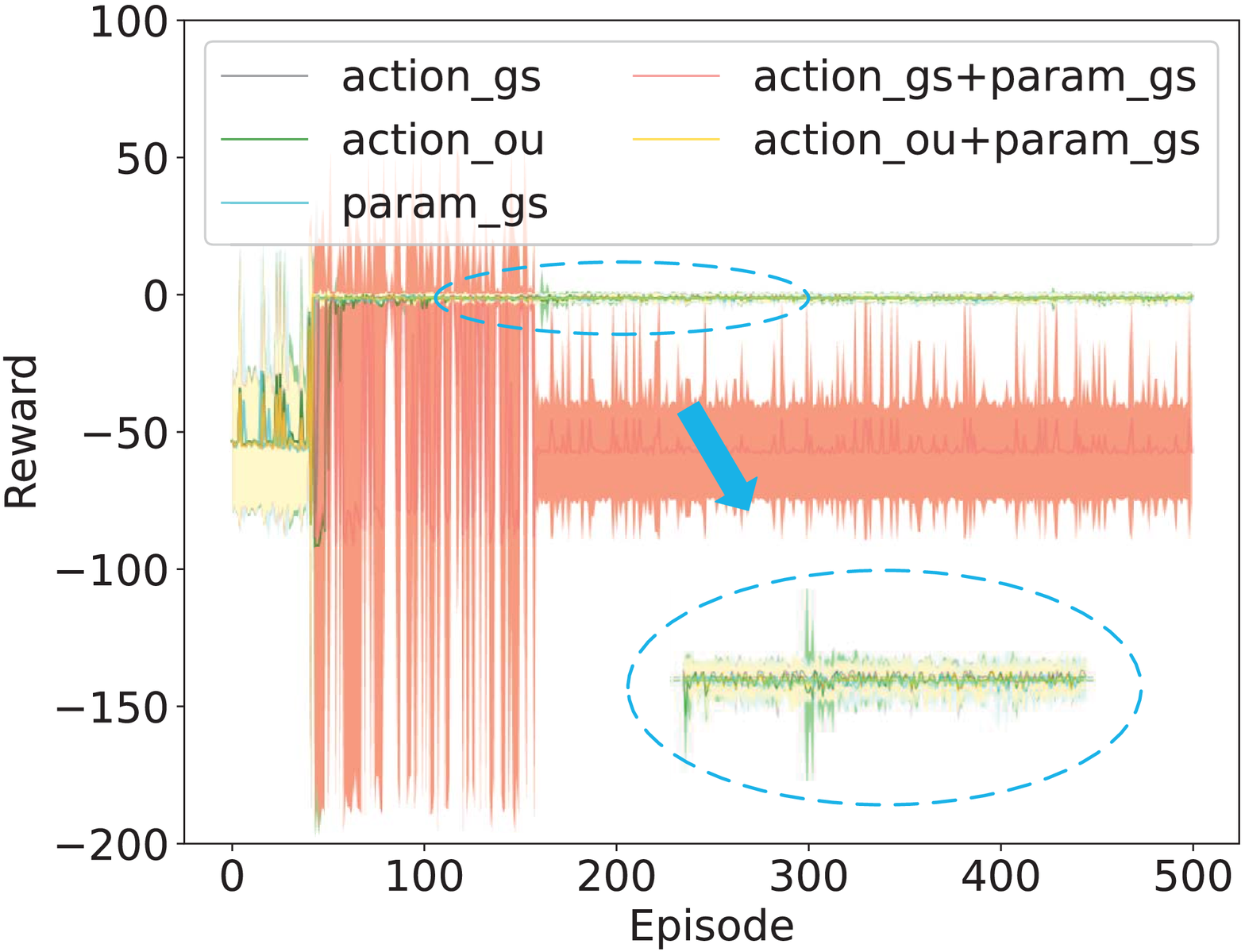}}\subfigure[Transferred networks]{
			\label{source_best}
			\includegraphics[width=0.32\linewidth]{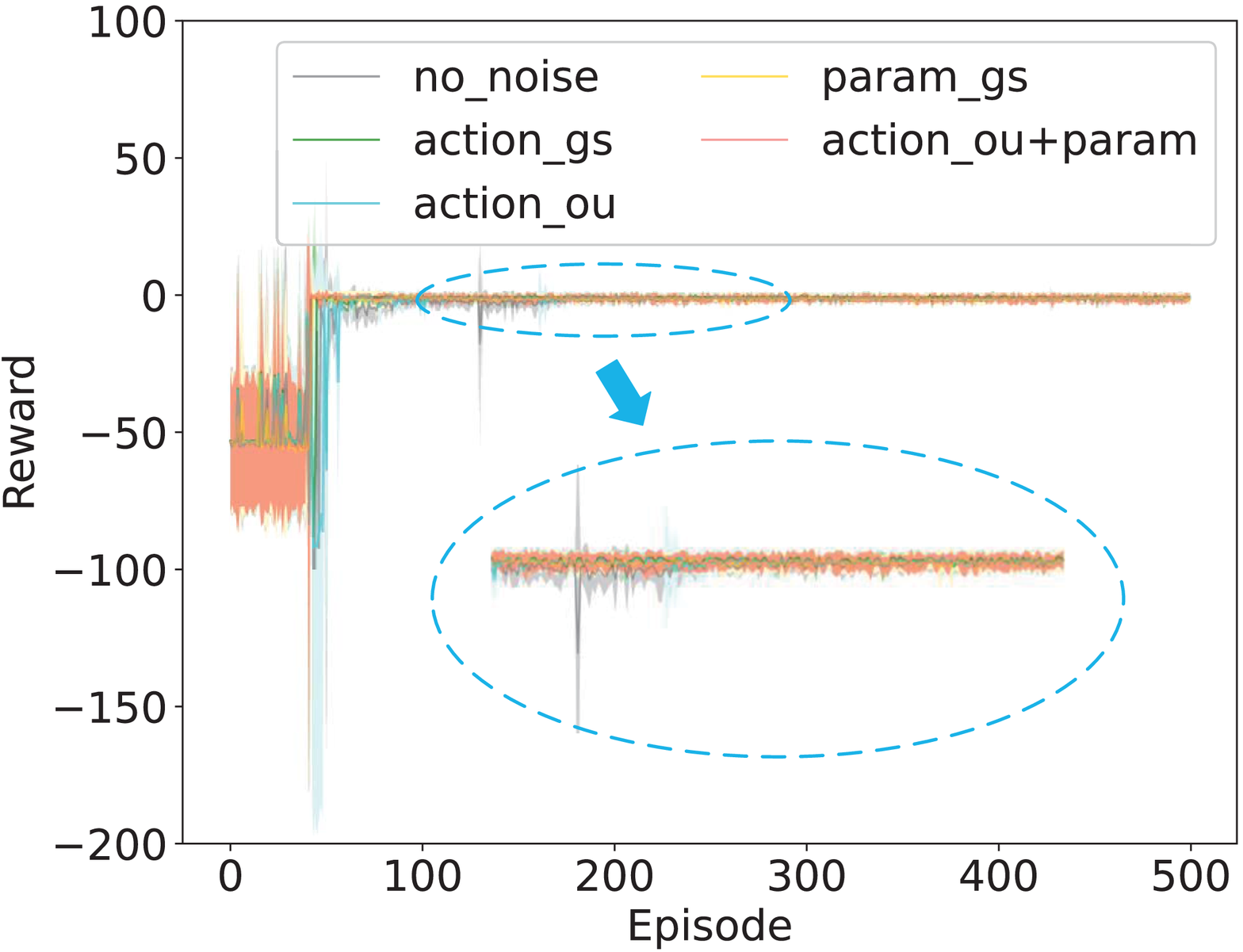}}
		\caption{Comparison of different exploration methods in the source domain.}
		\label{source}
	\end{figure*}
	
	As described in Fig.\ref{source}, Gaussian noise added in the action space, OU noise added in the action space, Gaussian noise added in the parameter space, and their mixture are used to explore in DDPG-based EMS. In Fig.\ref{source_gs}, different variance values $\sigma ^2$ of Gaussian noise added in the action space are set to 0.02, 0.03, 0.04, 0.05, 0.06, respectively. The reward fluctuates the most when the variance is set to 0.02. Only the network with variance 0.05 shows considerable oscillations once the trained weights converge. Thus, the network with variance 0.06 is the most stable. Similarly, Fig.\ref{source_ou} and \ref{source_param} illustrate that the network's variance of 0.09 with OU noise added in the action space and the network's variance of 0.03 with Gaussian noise added in the parameter space are the most stable, respectively. Results of other noise configurations, which converge to a local optimum or fluctuate too much, are not shown here. Besides, different effects of a single noise and a mixed noise are compared in Fig.\ref{source_mix}. Results using the Gaussian noise added in the action space and parameter space are very unstable, as shown by the yellow line. The noise of multiple Gaussian processes makes the agent tend to explore rather than exploit to a great extent, leading to more non-optimal actions with fluctuating reward values. The most stable is the Gaussian noise added in the action space, followed by the Gaussian noise added in both the action space and parameter space.
	
	Above all, the most stable network is the one that utilizes the Gaussian noise added in the action space with 0.06 variance to explore, followed by the network with the combination of OU noise added in the action space and Gaussian noise added in the parameter space to explore. Chosen networks which are transferred to a new domain are shown in Fig.\ref{source_best}.
	
	\subsection{Adaptation of Transfer Learning}
	
	Results of the transferred DDPG in the target domain using different exploration methods in EMS are discussed in this subsection. A new network is trained on the new driving cycle, learning from scratch or initializing the network parameters using prior ones.
	
	\begin{table*}[]
		\caption{Mean reward and Iteration Number of Target Network}
		\centering{
			\begin{tabular}{ccccc}
				\toprule
				\textbf{\begin{tabular}[c]{@{}c@{}}Exploration Method\\ (Target Network)\end{tabular}} & \textbf{\begin{tabular}[c]{@{}c@{}}Transferred Network Parameter\\ (Source Network)\end{tabular}} & \textbf{\begin{tabular}[c]{@{}c@{}}Mean Return\\ (First 50 Episodes)\end{tabular}} & \textbf{\begin{tabular}[c]{@{}c@{}}Mean Return\\ (Convergence Interval)\end{tabular}} & \textbf{\begin{tabular}[c]{@{}c@{}}Iteration\\ Number\end{tabular}} \\ \hline
				\multirow{5}{*}{No noise}                                                              & TFS                                                                                               & -21.6386                                                                           & -1.2101                                                                               & 30                                                                  \\
				& Gaussian\_AS                                                                                      & -0.9186                                                                            & \textbf{-1.0540}                                                                      & 20                                                                  \\
				& OU\_AS                                                                                            & -0.9451                                                                            & -1.2019                                                                               & 29                                                                  \\
				& Gaussian\_PS                                                                                      & \textbf{-0.9009}                                                                   & -1.4133                                                                               & \textbf{21}                                                         \\
				& APS                                                                                               & -2.0715                                                                            & -1.1878                                                                               & 29                                                                  \\ \cline{2-5} 
				\multirow{5}{*}{Gaussian noise added in the ation space}                               & TFS                                                                                               & -20.9453                                                                           & \textbf{-0.8766}                                                                      & 36                                                                  \\
				& Gaussian\_AS                                                                                      & -0.8780                                                                            & -0.8776                                                                               & 35                                                                  \\
				& OU\_AS                                                                                            & -0.9748                                                                            & -0.8908                                                                               & 30                                                                  \\
				& Gaussian\_PS                                                                                      & \textbf{-0.7987}                                                                   & -0.9071                                                                               & \textbf{25}                                                         \\
				& APS                                                                                               & -1.7628                                                                            & -0.9159                                                                               & 32                                                                  \\ \cline{2-5} 
				\multirow{5}{*}{OU noise added in the ation space}                                     & TFS                                                                                               & -25.3330                                                                           & -1.1527                                                                               & 36                                                                  \\
				& Gaussian\_AS                                                                                      & -1.1239                                                                            & -1.0585                                                                               & 24                                                                  \\
				& OU\_AS                                                                                            & -0.9035                                                                            & \textbf{-1.0417}                                                                      & \textbf{22}                                                         \\
				& Gaussian\_PS                                                                                      & \textbf{-0.8720}                                                                   & -1.1611                                                                               & 31                                                                  \\
				& APS                                                                                               & -2.3262                                                                            & -1.0474                                                                               & 23                                                                  \\ \cline{2-5} 
				\multirow{5}{*}{Gaussian noise added in the parameter space}                           & TFS                                                                                               & -21.6009                                                                           & -1.1523                                                                               & 35                                                                  \\
				& Gaussian\_AS                                                                                      & \textbf{-1.0250}                                                                   & \textbf{-1.1132}                                                                      & \textbf{21}                                                         \\
				& OU\_AS                                                                                            & -1.1109                                                                            & -1.1652                                                                               & 22                                                                  \\
				& Gaussian\_PS                                                                                      & -1.5669                                                                            & -1.2466                                                                               & 34                                                                  \\
				& APS                                                                                               & -4.4550                                                                            & -1.2734                                                                               & 30                                                                  \\ \cline{2-5} 
				\multirow{5}{*}{Noise added in ation space and parameter space}                        & TFS                                                                                               & -21.1288                                                                           & -1.0919                                                                               & 35                                                                  \\
				& Gaussian\_AS                                                                                      & -1.4819                                                                            & -1.0066                                                                               & 31                                                                  \\
				& OU\_AS                                                                                            & \textbf{-0.9408}                                                                   & -1.1473                                                                               & 27                                                                  \\
				& Gaussian\_PS                                                                                      & -13.8613                                                                           & \textbf{-0.9998}                                                                      & \textbf{24}                                                         \\
				& APS                                                                                               & -2.6120                                                                            & -1.1589                                                                               & 30                                                                  \\ \bottomrule
		\end{tabular}}
		\label{return}
	\end{table*}
	
	As shown in Table \ref{return}, following criteria are adopted to evaluate the adaptation of the transferable EMS \cite{r1}:
	
	\subsubsection{Jumpstart Performance (JP)}
	The initial performance of the agent. The mean reward of the first 50 episodes is used to evaluate it.
	
	\subsubsection{Asymptotic Performance (AP)}
	The ultimate performance of the agent. The mean reward of the convergence interval is used to evaluate it.
	
	\subsubsection{Time to Threshold (TT)}
	The learning time needed for the target agent to reach a certain performance threshold. The iteration number of convergence is used to evaluate it.
	
	\begin{figure*}[thpb]
		\centering
		\subfigure[No noise]{
			\label{no_noise}
			\includegraphics[width=0.32\linewidth]{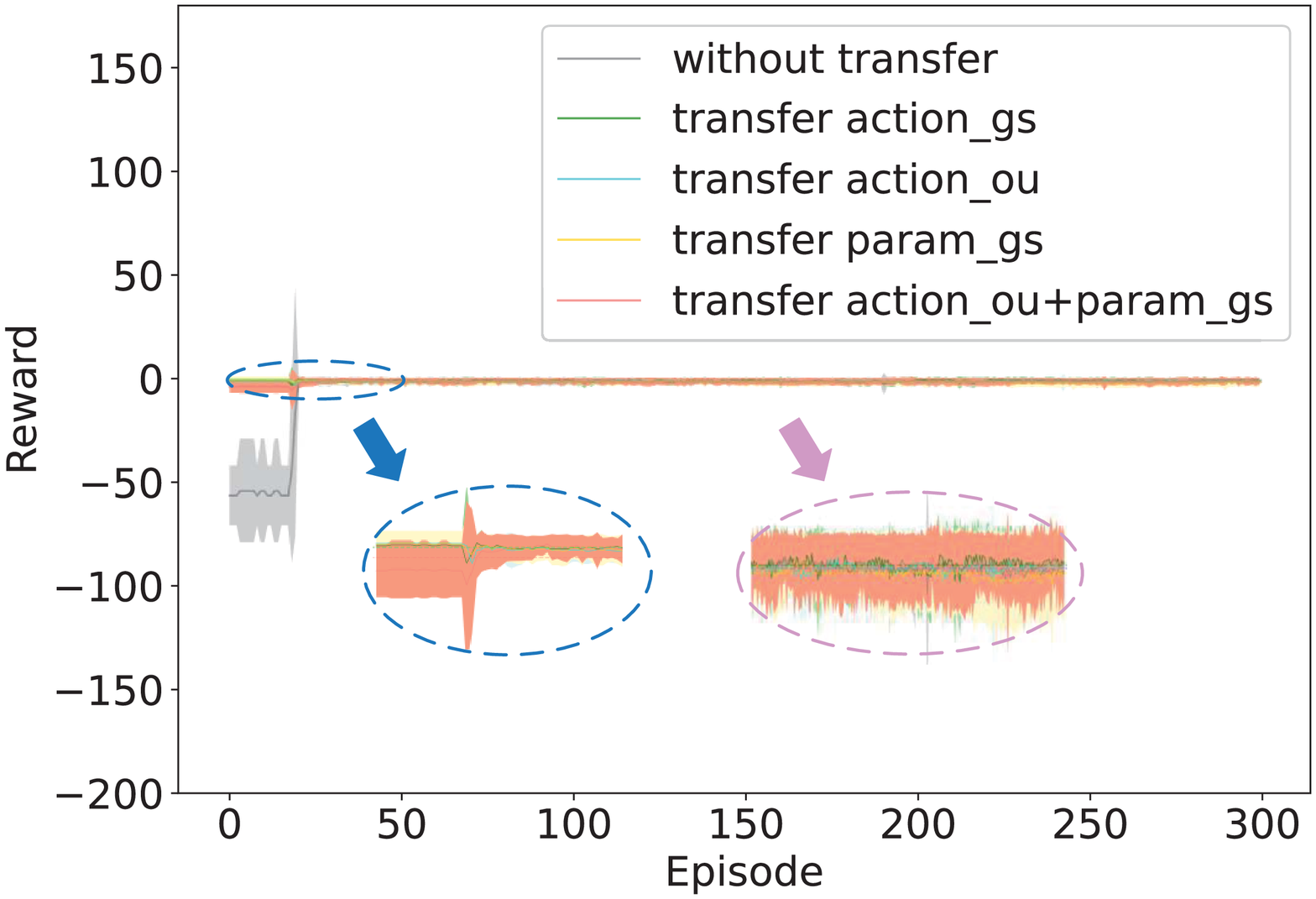}}
		\subfigure[Gaussian action space noise]{
			\label{gs}
			\includegraphics[width=0.32\linewidth]{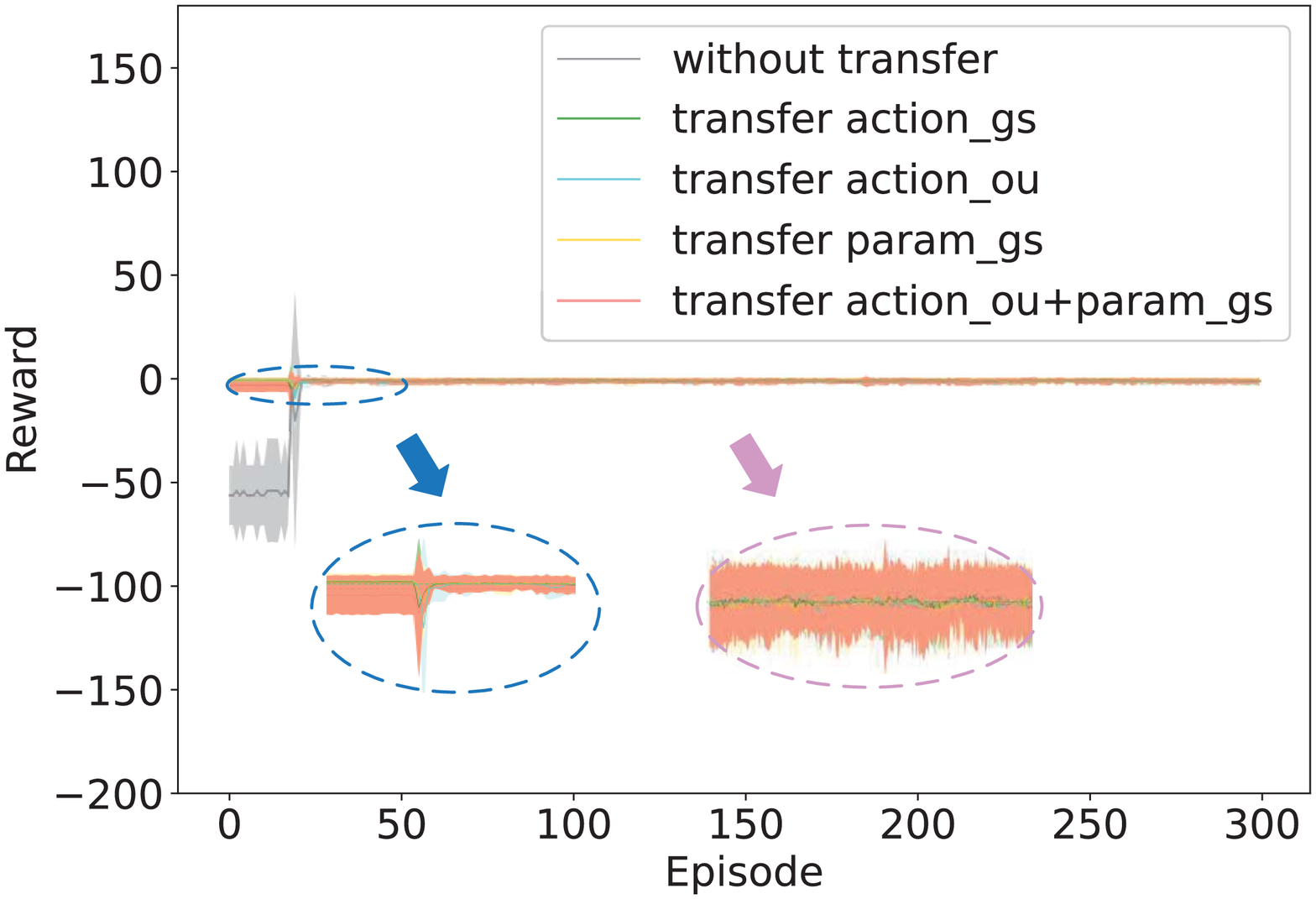}}\subfigure[OU action space noise]{
			\label{ou}
			\includegraphics[width=0.32\linewidth]{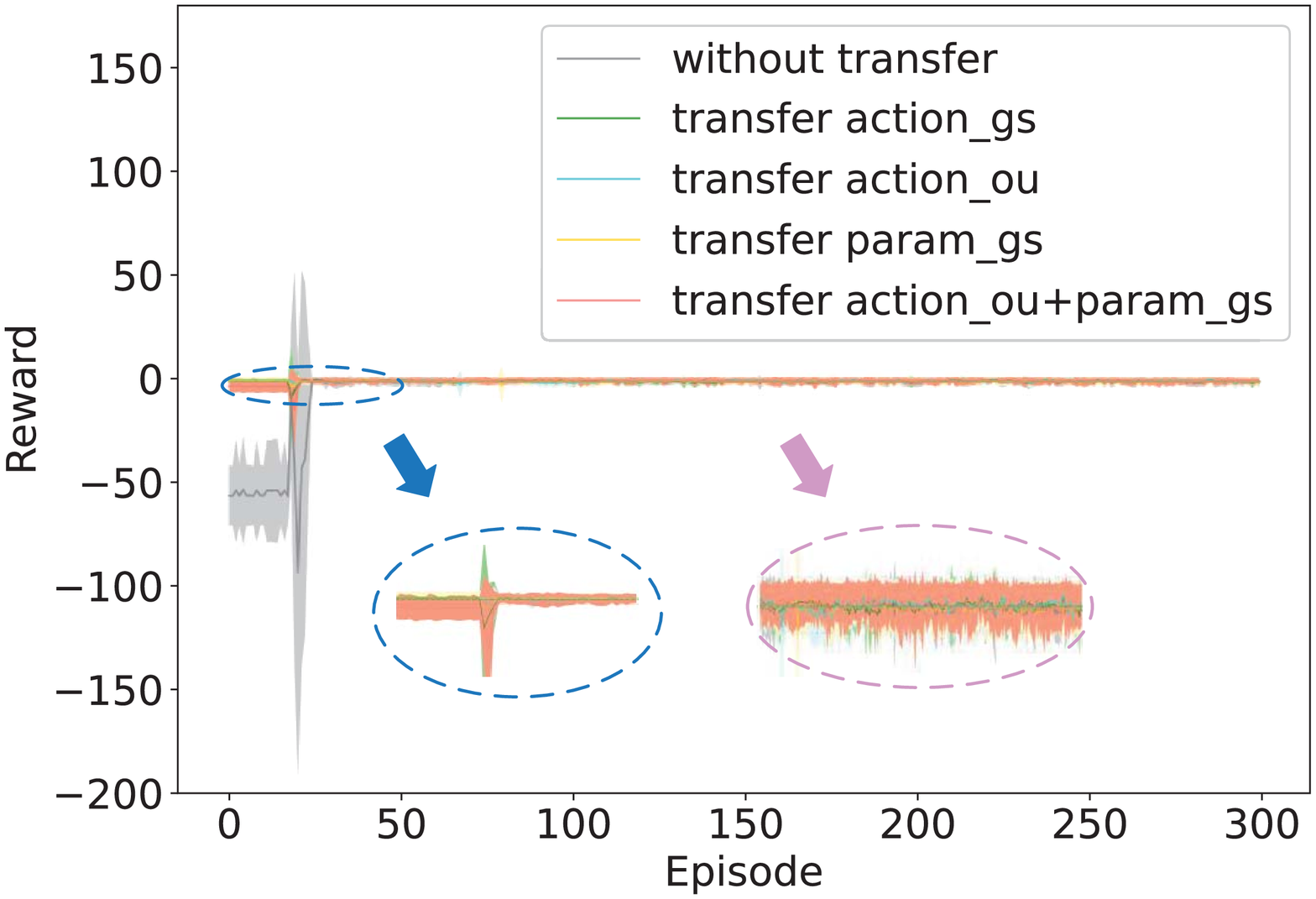}}
		\subfigure[Parameter space noise]{
			\label{param}
			\includegraphics[width=0.32\linewidth]{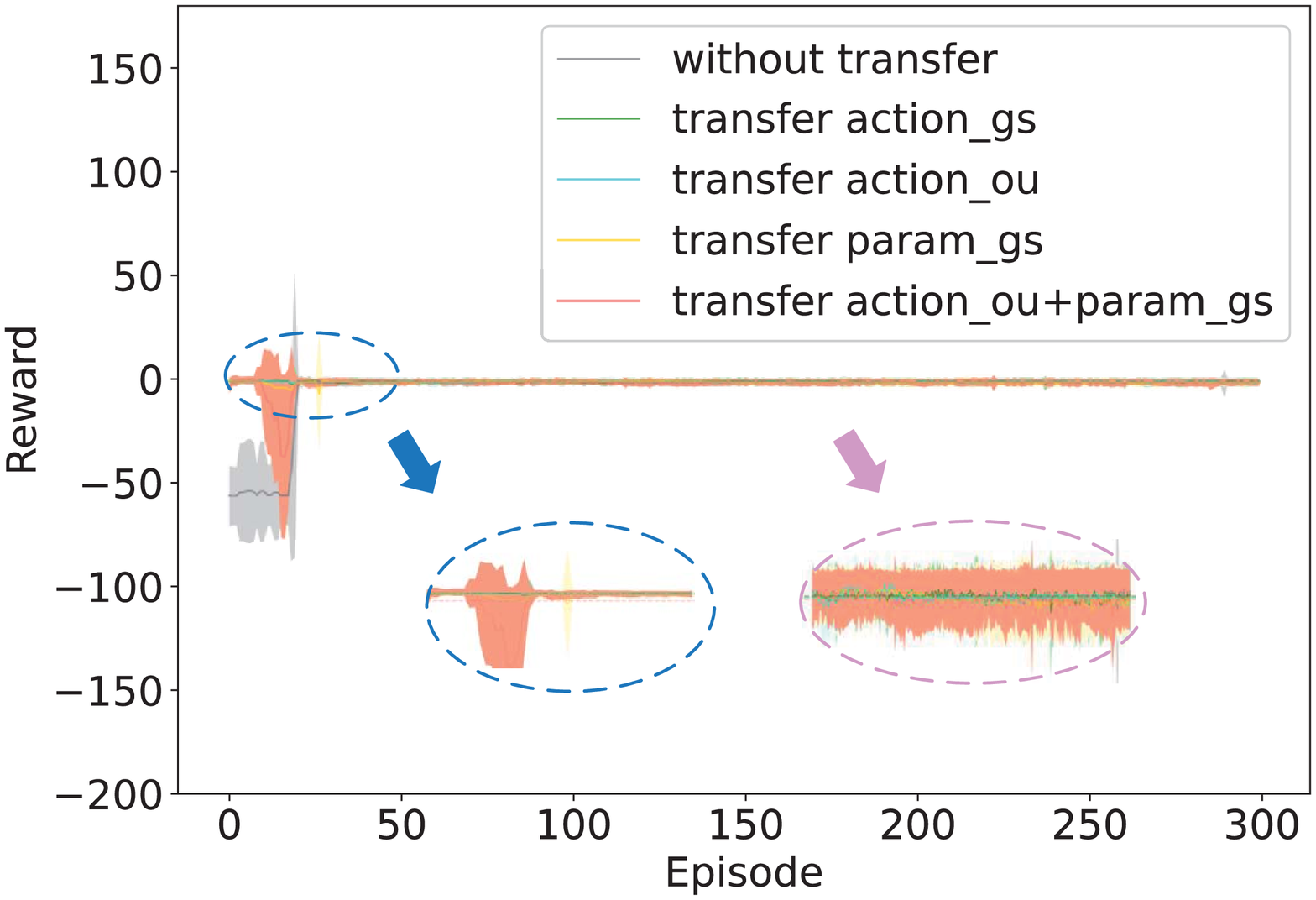}}\subfigure[Action space and parameter space noise]{
			\label{ou_param}
			\includegraphics[width=0.32\linewidth]{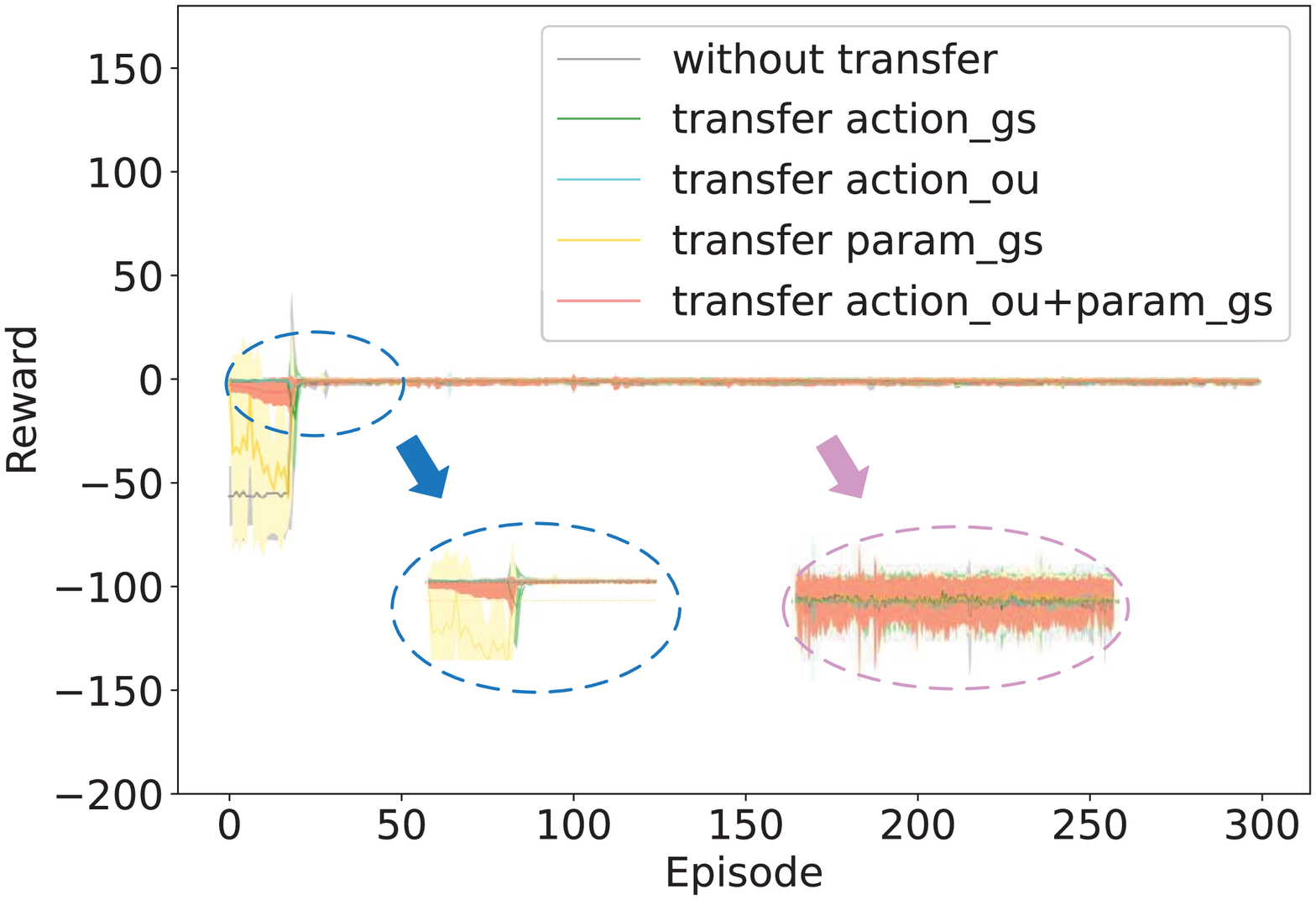}}
		\caption{Comparison of different exploration methods in the target domain.}
		\label{target}
	\end{figure*}

	In Table \ref{return}, TFS means the source network trained from scratch. Gaussian\_AS, OU\_AS, Gaussian\_PS and APS mean source networks with Gaussian noise added in the action space, OU noise added in the action space, Gaussian noise added in the parameter space, and both OU noise added in the action space and Gaussian noise added in the parameter space, respectively. While target networks use the same type of noise, the maximum reward value obtained from first 50 episodes is larger than the maximum convergence value in the convergence interval, which means that the DDPG network falls into a local optimum during the exploration process. Besides, the initialization of parameters of target networks with parameter space noise generally works well in terms of convergence speed and mean reward value, except for target networks using the mixture of action space and parameter space noise for exploration. After target networks have converged, convergence values of different exploration methods do not differ significantly.
	
	As shown in Fig.\ref{no_noise}, no noise is added on the target network. In first 50 episodes, the network which is trained from scratch (gray line) starts out with a small reward value, which means that its JP is poor. The green, blue, yellow, and orange lines represent parameters of networks with different exploration methods to initialize current target networks. Weights of these networks are perturbed by Gaussian noise added in the action space ($\sigma ^2 = 0.06$), OU noise added in the action space ($\sigma ^2 = 0.09$), Gaussian noise added in the parameter space ($\sigma ^2 = 0.03$), both OU noise added in the action space noise and Gaussian noise added in the parameter space to explore, respectively. The yellow line fluctuates the least. In the convergence interval (50 ~ 300 episodes in all target cases), the green line is the most stable, gray and blue lines are the second most unstable, and the most unstable one is the yellow line. In general, they do not differ much. This means that it is the most stable to initialize the target network with parameters of the original network with Gaussian action space noise.
	
	The Gaussian noise is added in the action space of target networks, of which the training process is shown in Fig.\ref{gs}. In first 50 episodes, the gray line fluctuates the most, and the orange line fluctuates the least, while the blue line fluctuates the least in the convergence interval. This indicates that the most stable approach is to initialize the target network with parameters of the original network with OU action space noise, and use Gaussian action space noise to explore.
	
	The action space noise with OU process is added on the target network, of which the training process is shown in Fig.\ref{ou}. The gray line has the most considerable fluctuation, followed by the orange line in first 50 episodes. The blue line has the slightest fluctuation. In the convergence interval, the most stable one is still the blue line, which indicates that it has a better learning effect to initialize the target network, which uses OU action space noise to explore, with parameters of the original network with OU action space noise.
	
	The Gaussian noise added in the parameter space is added on the target network, of which the training process is shown in Fig.\ref{param}. In first 50 episodes, the gray line has a small initial value because it does not have any prior knowledge. The orange line fluctuates the second most, which means that parameters of the network with action space noise and parameter space noise are not suitable for initializing the new target network with parameter space noise. In the convergence interval, the most stable one is still the blue line, followed by the green line, and the most unstable one is the yellow line. The blue line is the most stable one throughout the training process.
	
	Both the action space noise and the parameter space noise are added on the target network, of which the training process is shown in Fig.\ref{ou_param}. The gray line has the smallest initial value, while the yellow line has the largest fluctuation in first 50 episodes. It means that parameters of the network with parameter space noise are not suitable for initializing the target network with action space noise and parameter space noise. In the convergence interval, the fluctuation of each method is negligible.
	
	Combining results expressed in Fig.\ref{source} and Fig.\ref{target}, the network with the action space noise works best for DDPG-based EMS. For transfer learning, the network with the parameter space noise is the most stable, while the network with multiple noises of action space and parameter space has poor initial performance.
	
	\section{Conclusion}
	
	In this paper, choosing an optimal and efficient EMS is formulated as a deep reinforcement learning-based transfer learning problem. We compared different exploration approaches for deep reinforcement learning and transfer learning to find out the best energy management strategy. Effects of action space noise and parameter space noise which are added to the DDPG algorithm, are presented. Experimental results show that the method of parameter space noise exploration works best for DDPG-based transferable EMS.
	
	Despite these encouraging comparison results, there are several avenues for future research. First, we want to investigate other exploration modalities, not only adding noise for action selecting. We furthermore plan to compare different exploration methods based on more robust DRL algorithms, such as twin delayed DDPG and soft actor-critic.
	
	\addtolength{\textheight}{-12cm}   
	



	
	\bibliographystyle{IEEEtran}
	\bibliography{ref}

\end{document}